\pdfoutput=1

\documentclass[11pt]{article}

\usepackage[]{acl}

\usepackage{times}
\usepackage{latexsym}
\usepackage{graphicx}
\usepackage{booktabs}
\usepackage{multirow}

\usepackage[T1]{fontenc}
\usepackage[utf8]{inputenc}
\usepackage{microtype}
\usepackage{amsmath}
\DeclareMathOperator*{\argmax}{arg\,max}

\title{It's All Relative! -- A Synthetic Query Generation Approach for \\ Improving Zero-Shot Relevance Prediction   }

\author{Aditi Chaudhary \And
  Karthik Raman \And
  Michael Bendersky \\
  {\hspace{-25em} Google Research, USA} \\
  {\hspace{-25em}\texttt{\{aditichaud, karthikraman, bemike\}@google.com}}}

\begin{document}
\maketitle
\begin{abstract}
Recent developments in large language models (LLMs) have shown promise in their ability to generate synthetic query-document pairs by prompting with as few as 8 demonstrations \cite{dai2022promptagator}.
This has enabled building better IR models, especially for tasks with no training data readily available.
Typically, such synthetic query generation (QGen) approaches condition on an input context (e.g. a text document) and generate a query relevant to that context, or condition the QGen model additionally on the relevance label (e.g. relevant vs irrelevant) to generate queries across relevance buckets.
However, we find that such QGen approaches are sub-optimal as they require the model to reason about the desired label and the input from  a handful of examples.
~In this work, we propose to reduce this burden of LLMs by \emph{generating queries simultaneously for different labels}.  
We hypothesize that instead of asking the model to generate, say, an irrelevant query given an input context, asking the model to generate an irrelevant query \emph{relative to a relevant query} is a much simpler task setup for the model to reason about. 
Extensive experimentation across seven IR datasets shows that synthetic queries generated in such a fashion translates to a better downstream performance, suggesting that the generated queries are indeed of higher quality.

\end{abstract}

\section{Introduction}
Predicting how relevant a query is to a document has many real-world applications, ranging from  e-commerce product search \cite{reddy2022shopping} to scientific literature search \cite{wang2020cord,cohan-etal-2020-specter}.
Collecting data for such diverse use-cases is often infeasible, especially when new topics (e.g. COVID-19) or  products (e.g. iPhone 15) are introduced.
Given the importance of in-domain data \cite{gururangan-etal-2020-dont}, the question arises: if data collection is challenging, what about data generation?

\begin{figure*}
    \centering
    \includegraphics[width=\textwidth]{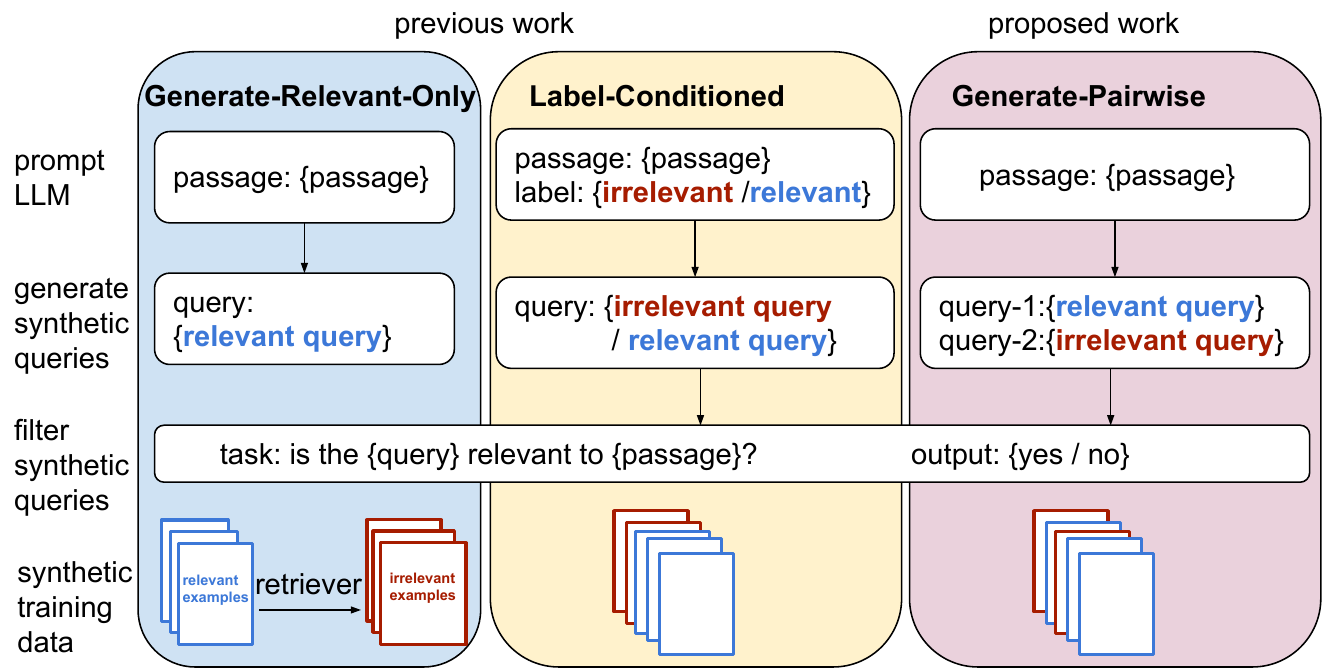}
    \caption{An overview of our proposed pairwise query generation approach (\textsc{Generate-Pairwise}), where we prompt the model to generate both the relevant and the irrelevant query at the same time, forcing the model to condition the irrelevant query generation on the previous generated query.
    We compare the proposed approach with previous QGen approaches, \textsc{Generate-Relevant-Only}, where only relevant queries are generated followed by a retriever to retrieve irrelevant examples and, \textsc{Label-Conditioned}, where the query generation is conditioned on one relevance label, at a time, to generate a query.
    Next, the generated queries are filtered automatically to ensure self-consistency. and the resulting synthetic data is used for downstream training.}
    \label{fig:overview}
\end{figure*}
Recently, Large Language Models (LLMs) (GPT-3:\citet{DBLP:conf/nips/BrownMRSKDNSSAA20}, PaLM: \citet{DBLP:journals/corr/abs-2305-10403}) have been applied to automatically generate task-specific data which can be used to train downstream models.
For example, Promptagator \cite{dai2022promptagator} and InPars \cite{bonifacio2022inpars}, use as few as 8 task-specific demonstrations and prompt the LLM to generate synthetic queries for new documents.
These demonstrations are pairs of input-output examples where the input is a document text and the expected output is a query relevant to that context.
However, this process only generates ``relevant'' synthetic queries.
To create the corresponding negative or ``irrelevant''  query-document pairs, a retriever is used to retrieve documents for each synthetic query, from which the hard negatives  are constructed.
Finally, a task-specific downstream model is  trained using this synthetic training data.
\citet{chaudhary2023exploring} forgo the retriever step, and instead generate queries across different relevance buckets (e.g. relevant and irrelevant) by providing the relevance label along with the document text as the context.
Although, they find conditioning query generation (QGen) on the different relevance labels does outperform approaches such as Promptagator, where the QGen is only conditioned on the document context, the downstream model trained on the task-specific synthetic data is outperformed by the traditional transfer learning model, where a model is trained on a related dataset and directly applied to the target task.
This suggests that there is significant room of improvement for QGen models.

In this work, we propose a novel few-shot QGen approach where we prompt the model to \emph{generate queries across different relevance labels relative to each other}.
For example, given an input document  and two relevance labels (relevant and irrelevant), the model is required to generate both a relevant query and an irrelevant query. 
By forcing the model to generate an irrelevant query after generating a relevant query, forces the model to condition on the previously generated query along with the document context.
The process is illustrated in \autoref{fig:overview}.
We hypothesize this is a much more simpler task setup for the model in comparison to generating, say, an irrelevant query from only the document context or the label alone.
This is inspired from the observation that both humans and LLMs  find the task of comparing two things with respect to each other a simpler task, as opposed to judging things in isolation \cite{christiano2017deep, qin2023large}. 
To evaluate our hypothesis, we conduct experiments with seven IR datasets from the BEIR benchmark \cite{thakur2021beir} that have no training data. 
Therefore, we generate synthetic data for each dataset by prompting the LLM with only 10 examples from MS-MARCO \cite{msmarco} and train a downstream model on the generated data.
We use the downstream task performance as a signal to evaluate the effectiveness of the QGen model and our key observations are:
\begin{itemize}
    \item Generating queries relative to each other leads to an improved downstream score -- (+8.6 NDCG@10) avg. improvement over existing QGen approaches for five out of six datasets. (\autoref{tab:results})
    \item For three datasets, the downstream model trained only on our generated queries even outperforms the strong skyline of transfer learning model which is fine-tuned on $\sim 6M$ MS-MARCO examples   and directly applied to the target task. For two datasets, the best QGen model is only 3 points behind this strong skyline model, suggesting the generated data is almost at par with human-labeled data. 
    \item A case study on fine-grained relevance prediction shows that generating pairwise queries is also well-suited to generate nuanced queries across multiple relevance labels (\autoref{tab:wands}).
\end{itemize}

\section{Synthetic Query Generation}
We follow  Promptagator's few-shot query generation (QGen) setup, which
 comprises of three steps, namely, prompt-based query generation, query filtration, and, downstream model training.
Below, we describe each step in more detail.

\subsection{Background: Prompt-Based QGen} \label{sec:qgen}
The first step of the QGen process generates task-specific query-document pairs using an instruction prompt $p$ and the target task $T$'s document corpus $\mathcal{D_T}$.
Existing approaches such as Promptagator and InPars generate relevant query-document pairs.
Specifically, Promptagator use $k$ task-specific relevant query-document pairs as the instruction prompt $p_T\!=\!\{(d_i, q_i)^k, d_t\}$, where the document $d_i$ is relevant to the query $q_i$, and $d_t$ denotes the new document for which we want to generate the query.
The prompt $p_T$ is then run on the target task corpus $\mathcal{D}_T$ to generate a large relevant query-document set.
\citet{chaudhary2023exploring} condition the QGen model on the relevance label as well, which allows them to generate query-document pairs across different relevance labels ($l \in L$), where $L$ is the set of all relevance labels.
In this case, the instruction prompt $p_T\!=\!\{(l_i, d_i, q_i)^k, (l_t, d_t)\}$ comprises of $k$ query-document-label triplets, where for the document $d_i$ the query $q_i$'s relevance label is $l_i$.
For instance, in the case of binary relevance labels, the prompt examples will contain both relevant  and irrelevant query-document pairs.
Now, to generate a new query, the prompt takes the desired relevance label ($l_t$) along with the new document ($d_t$), generating a new query whose relevance level with the document is implied by the relevance label.

\subsection{Proposed Work: Pairwise Query Generation} \label{sec:proposed}
The above approaches, both the label-conditioned \cite{chaudhary2023exploring} and the non-label conditioned Promptagator \cite{dai2022promptagator}, generate queries in isolation of other queries.
As motivated in the introduction, comparing two outputs with respect to each other is an easier setup as opposed to evaluating an output in isolation \cite{christiano2017deep, qin2023large}.
This makes us wonder whether formulating the QGen task in a similar fashion would lead to high-quality synthetic data.

Specifically, we propose pairwise query generation, where given an input document, the model is required to generate both the relevant and the irrelevant query.
Like before, we  construct an instruction prompt $p$ using $k$ input-output examples, where the input is a document $d$ and and the output is a pair of queries $(q_{\text{rel}}, q_{\text{irrel}})$ where the first query $q_{\text{rel}}$ is relevant to the $d$ while the second query $q_{\text{irrel}}$ is irrelevant.
The pairwise prompt, therefore, is as follows:
$p\!=\!\{(d_i, q_\text{rel}, q_\text{irrel})^k, d_t\}$.
Given a new document, the model is expected to generate both the relevant and the irrelevant queries at the same time, as shown in \autoref{tab:prompt_formats}.
By forcing the model to generate both queries, the generation of the irrelevant query is conditioned on the previously generated relevant query along with the input context.
Note, the generated irrelevant query will be thematically similar to the input document and not completely unrelated.
In other words, these queries are ``hard negatives'', which previous works \cite{gao2021complement, karpukhin2020dense, guu2020retrieval} have shown to be better for training downstream IR models rather than using randomly sampled or in-batch negatives.

\subsection{Query Filtration} \label{sec:filter}
After generating the synthetic query set, round-trip consistency filter \cite{alberti-etal-2019-synthetic} is applied to remove noisy queries.
For example, Promptagator use a retriever trained on the synthetic data to check whether the generated query is relevant to the document from which it was generated, and find that including the round-trip filtering is important for improving the query quality.
Given the superiority of LLMs, we prompt the LLM again, but this time to predict whether the generated synthetic query $q_{\text{syn}}$ is relevant or not to the document $d_t$ from which it was generated.
Specifically, the prompt comprises of $k$ query-document example pairs across both relevance labels:
$p_{\text{filter}}\!=\!\{(q_i, d_i, l_i)^k, (q_{\text{syn}}, d_t)\}$, where $l \in L\!=\!\{\text{relevant}, \text{irrelevant}\}$.
We apply the $p_{\text{filter}}$ on the synthetic query-document pair $(q_{\text{syn}}, d_t)$ and compute the log-likelihood to score each output label $L$, given the prompt.
Then, we select the label with the highest score, as shown below:
$l_{\text{pred}} = \argmax_{L} \text{score}(l, p_{\text{filter}})$ where $p$ refers to the filtration prompt and \emph{score} is the log-likelihood.
We only retain those queries whose predicted label $l_{\text{pred}}$ matches the label for which the query was generated.
For instance, in our proposed pairwise QGen, the first query will be retained only if it is rated as relevant and, similarly, the second query if it is rated as not relevant.
In addition to the round-trip filtering, we also remove duplicate queries, i.e. queries that got labeled with different relevance labels for the same document, following \citet{chaudhary2023exploring}.

\subsection{Downstream Model Training} \label{sec:downstream}
Finally, a downstream model is trained on the filtered synthetic data.
For Promptagator, this entails training a task-specific retriever and a re-ranker on the synthetic query-document pairs generated for that task, as their end-task is to  improve document retrieval across diverse tasks.
Similar to \citet{chaudhary2023exploring} we train a relevance prediction model for each of the target tasks, since our goal is to improve zero-shot relevance prediction for a new task where there is no training data.
Note that through our proposed method, the model is able to generate training examples for both relevant and irrelevant labels, alleviating the need of running the additional step of an retriever, as done by \citet{dai2022promptagator} and \citet{bonifacio2022inpars}.

\section{Experimental Setup} \label{sec:setup}
Our goal is to improve relevance prediction  for tasks with no training data, for which we generate sufficient synthetic data and train a downstream model.
Therefore, we experiment with seven IR tasks from the popular BEIR benchmark \cite{thakur2021beir}, under zero-shot settings, where we generate query-document examples for each task.
We aim to cover a balanced mixture of tasks covering diverse task categories, domains, and number of relevance labels.\footnote{More details on the data in \autoref{sec:app_data}}.
For all tasks, we construct the prompt using a fixed set of examples from the MS-MARCO dataset \cite{msmarco}, which is a general-purpose retrieval dataset covering a variety of domains. Below we summarize all our datasets. Note, all datasets are based in English.

\subsection{Data}
\paragraph{MS-MARCO} released by \citet{msmarco} is constructed from Bing search logs comprising of general purpose web query-document pairs.
It comprises of 530,000 queries and 8 million passages, covering two relevance labels, namely, relevant and irrelevant.

\paragraph{TREC-COVID} \cite{voorhees2021trec} consists of scientific query-document pairs based on the COVID-19 pandemic.
It comprises of 171,332 passages, with a test set of 35,480 examples across three relevance labels, namely, relevant ($2.0$), somewhat relevant ($1.0$) and irrelevant ($0.0$).
The BEIR task category for this dataset is Bio-medical IR.

\paragraph{FIQA-2018} \cite{maia201818} (Task-2) is  opinion-based question answering (QA) task. BEIR mines 57,638 financial articles as the document corpus, with a test set comprising of 1,706 relevant query-document examples.
The BEIR task category for this dataset is open-domain QA.

\paragraph{Touch\'e20} \cite{bondarenko2020overview} (Task 1) is a conversational argument retrieval task, with 382,545 documents constructed from the conclusion and premise of arguments.
The test set comprises of 2,099 query-document examples, covering three labels, namely, relevant ($2.0$), somewhat relevant ($1.0$) and irrelevant ($0.0$).
The BEIR task category for this dataset is Argument Retrieval.

\paragraph{DBPedia} \cite{hasibi2017dbpedia} is an entity retrieval dataset, with 4,635,922 DBPedia English articles.
The test set has 40,724 query-document examples, covering three labels, namely, relevant ($2.0$), somewhat relevant ($1.0$) and irrelevant ($0.0$).
The BEIR task category is Entity Retrieval.

\paragraph{FEVER} \cite{thorne2018fever} is a fact-verification task, with 123,142 sentence claims as queries and 5,416,568 Wikipedia abstracts as the document corpus. The test set has 1,499 relevant query-document examples.
The BEIR task category for this dataset is Fact Checking.

\paragraph{Climate-FEVER} \cite{diggelmann2020climate} is also a fact-verification task, with 5,416,593 Wikipedia abstracts, but with 1,535 real-world climate claims as queries. The test set has 1,344 relevant query-document examples. The BEIR task category for this dataset is Fact Checking.

\subsection{Model}
As described above, our QGen process involves three steps: query generation, query filtration, and downstream model training. 

\paragraph{QGen Model}
For both the query generation and query filtration, we use PaLM-2 (M) \cite{DBLP:journals/corr/abs-2305-10403}, a multilingual LLM pretrained on a variety of corpora such as web documents, books, code, mathematics and conversational data, to name a few.
For all tasks, we select 10 query-document examples from the MS-MARCO to curate the prompt (more details on prompt construction in \autoref{sec:baseline} and \autoref{sec:proposedmodel}).
Given that for some datasets the corpus of documents is quite large, we randomly sample 50,000 documents for each task to apply the QGen model, except for TREC-COVID where we use all the test document corpus comprising of 171k passages in order to measure the effect of using the entire corpus.
We use temperature sampling to generate the outputs, with temperature of 0.6 and beam size of 2.
In \autoref{tab:relevant_qgen}, \autoref{tab:pairwise_qgen}, \autoref{tab:labelcond_qgen}, and \autoref{tab:iterative_qgen}, we report the number of queries generated at each step of the QGen process.

\paragraph{Downstream Model}
To evaluate the quality of the generated queries, we train a  pointwise encoder model using mt5-XXL \cite{xue-etal-2021-mt5} for the task of relevance prediction and score the test query-document examples. 
We use all of the filtered synthetic data as our training split.
As validation split, we sample 5k examples (2,500 for each binary relevance label) from MS-MARCO, and use the test split of the target task to report the final metric.
Similar to \citet{chaudhary2023exploring}, we use the NDCG metric to report the downstream model performance.\footnote{As explained in Section 4.3 of \citet{chaudhary2023exploring}, using NDCG over accuracy avoids the need for mapping labels across different datasets, as many test datasets do not have the same label set as the train dataset i.e. MSMARCO; we refer the reader to the paper for more details.}
NDCG measures the ranking performance i.e. whether the model is able to rank true relevant documents higher than any other document.
For some datasets such as TREC-COVID, Touch\'e, and DBPedia, the test split already contains examples of not relevant and partially relevant query-document examples, in addition to the relevant examples.
However, for the other three datasets, the test split only provides relevant examples, assuming every other document to be irrelevant.
Therefore, for such datasets, following \citet{zhuang2022rankt5} and \citet{dai2022promptagator}, we use BM25 to retrieve top-20 documents for each test query and combine that with gold-annotated relevant examples, which allows us to compute the NDCG scores.
We train the mt5-XXL model to run for at least 1 epoch of the training data, using a learning rate of $5e-05$, batch size of 64, input sequence length of 512, and Adafactor optimzer.
Note that we use a  smaller model  for our downstream task as compared to our query generation model, as a smaller downstream model is more compute and time efficient, which is extremely important for real-world downstream application, where the inference needs to be run several times and often on large test sets.

\subsection{Baselines} \label{sec:baseline}

\paragraph{Transfer Learning}
Fine-tuning a model on a related dataset and applying it as-is on a new target domain, has been a simple but effective strategy to improve zero-shot performance.
Specifically, for all of the BEIR datasets, we fine-tune a mt5-XXL model on the general purpose MS-MARCO dataset.
We sample 6 million query-document examples, equally distributed across both relevant and irrelevant labels.
We treat this model as a \emph{skyline} given that this model has access to all of the MS-MARCO training data while for all the below QGen models only 10 selected examples are used.

\paragraph{Generate-Relevant-Only}
We prompt the LLM to generate relevant-only queries, similar to Promptagator \cite{dai2022promptagator}.
In this case, the prompt consists of 10 MS-MARCO examples of relevant query-document pairs.
As mentioned above, we use beam-size$\!=\!2$, which means, in this case, two relevant queries are generated for each document.
Next, a filtration prompt is applied to remove queries that are not relevant to the document from which they were generated.
We then use a T5-based retriever \cite{ni2021large,t5x-ret-sent} to retrieve hard negatives i.e. one irrelevant document for each synthetic query.
This setup is similar to Promptagator with the difference that we use the latest PaLM-2 model as our LLM, while Promptagator use the older Flan-PaLM model \cite{wei2021finetuned}.
Since our main goal is to evaluate the quality of different QGen models, we use a general-purpose retriever while Promptagator train a task-specific retriever.
Furthermore, Promptagator use 8 task-specific examples of relevant query-document pairs, while we use 10 examples from MS-MARCO, keeping true to the zero-shot assumption.

\paragraph{Label-Conditioned}
Following \citet{chaudhary2023exploring}, we prompt the model to generate a query by using the required relevance label and the document as context.
Specifically, the prompt includes 5 relevant query-document examples and 5 irrelevant query-document examples from MS-MARCO.
For a new task document, we then generate two synthetic queries for both relevant and irrelevant label each. 
Next, we prompt the LLM with the generated query and document to predict the relevance label and filter those queries whose desired label does not match the predicted label.
Since the model already generates irrelevant examples, we directly train the downstream model on the above data, without running the retriever.

\subsection{Proposed Models} \label{sec:proposedmodel}
We experiment with the following two model variants under the pairwise query generation approach.
\begin{table*}
\resizebox{\textwidth}{!}{
\begin{tabular}{llllllll}
\toprule
& & \multicolumn{3}{c|}{Gold Negatives} & \multicolumn{3}{c}{Gold + BM25 Negatives}\\
\midrule
& \textbf{Model} & \textbf{TREC-COVID} & \textbf{Touch\'e} & \textbf{DBPedia} & \textbf{Climate-Fever} & \textbf{FIQA} & \textbf{FEVER} \\

& & \textbf{(all)} & \textbf{(50k)} & \textbf{(50k)} & \textbf{(50k)} & \textbf{(50k)} & \textbf{(50k)} \\
\midrule
Skyline & Transfer Learning & 	0.7615 &	0.7714 &	\underline{0.6181} &	0.5129	& \underline{0.6781} &	\underline{0.9041} \\
\midrule
Baseline & Generate-Relevant-Only & 0.4145 &	0.6512 &	0.5471 &	0.4704 &	0.6027 &	\textbf{0.7915}\\
& Label-Conditioned & 0.7765 &	0.7466 &	0.5152 &	0.4888 &	0.4005 &	0.7032 \\
\midrule
Ours & Generate-Pairwise & 0.7831 &	\textbf{0.7881} &	\textbf{0.5631} &	\textbf{0.5562} &	0.4769 &	0.7235 \\
& Iterative-Pairwise & \textbf{0.8004}&	0.7762	& 0.5557	& 0.4972	& \textbf{0.6584} &	0.7054 \\
\bottomrule
\end{tabular}
}
\caption{We compare the NDCG@10 metric on the test split of all tasks (higher the better), here the mt5-XXL is our downstream model trained on the task-specific synthetic data. 
`Gold Negatives' refer to those datasets where the negative  examples are also provided by the BEIR benchmark, while `Gold+BM25 Negatives' refer to the datasets where no gold negatives are provided in the test file. 
For such datasets, BM25 is used to retrieve top-20 documents which are combined with the gold-annotated positive examples (i.e. relevant documents), over which the NDCG@10 is reported.
The number in the brackets below each dataset refers to number of documents used for query generation.
Overall best scores are \underline{underlined} and the best scores among the QGen models are \textbf{highlighted}.
}
\label{tab:results}
\end{table*}
\paragraph{Generate-Pairwise}
We prompt the model to generate both the relevant and the irrelevant query for a given document, at the same time.
Specifically, we prompt the model with the same 10 relevant query-document examples as above, but for each relevant query-document pair, we additionally also add an irrelevant query, as shown in \autoref{tab:prompt_formats}.
We generate two outputs for each document, which in this case means two relevant and two irrelevant queries.
Next, we filter the queries using the filtration prompt and train the downstream model.
Similar to the \textsc{Label-Conditioned} model, the retriever step is not applied as both relevant and irrelevant examples are generated.

\paragraph{Iterative-Pairwise}
Similar to the \textsc{Generate-Relevant-Only} model, we first generate relevant-only queries for each document and filter queries that are rated as not relevant.
Next, we prompt the LLM again with the same document context along with the filtered relevant query and prompt the model to generate an irrelevant query, using the same prompt as \textsc{Generate-Pairwise} model.
Then, we apply the filtration prompt on the generated irrelevant queries, and remove queries that are rated as relevant.
This is similar in principle to the above variant, where we still condition the generation of the irrelevant query on the previous generated relevant query.
Like the above model, for every QGen step we generate two outputs.

\section{Results and Discussion} \label{sec:results}
We report our main results in \autoref{tab:results}.
First, we find that among all query generation approaches, our proposed pairwise query generation results in the best downstream performance for five out of the six tasks.
We observe that simply conditioning the QGen model on the relevance label (\textsc{Label-Conditioned}) in addition to the document context already outperforms the \textsc{Generate-Relevant-Only} model by +2 NDCG@10 points (avg. across all tasks).
This is in-line with \citet{chaudhary2023exploring} who also find that conditioning on labels  helps  capture fine-grained nuance.

The proposed pairwise models further improves upon the \textsc{Label-Conditioned} models by +7 NDCG@10 points (avg.) across all tasks, suggesting that conditioning on a previous generated query provides an even stronger signal to the LLM to directly generate ranked query-document pairs, which aligns with our downstream task.
Interestingly, we find that for three of the datasets (TREC-COVID, Touch\'e, and Climate-FEVER), the models solely trained on synthetic data even outperforms the transfer learning skyline, where the entire MS-MARCO data was used for fine-tuning.
This is unlike \citet{chaudhary2023exploring}, where  none of the QGen approaches outperformed the simple but strong transfer learning skyline.
What makes this result interesting is that in all of the QGen models, only 10 MS-MARCO examples are used, highlighting again that a) \emph{quality matters over quantity} i.e. LLMs can learn to generate high-quality data given a good set of exemplars, and b) even noisy in-domain training data is effective than out-of-domain gold annotated data.

However, we do find that for the FEVER task, \textsc{Generate-Pairwise} is not the best performing model.
This is probably because there is a mis-match between the irrelevant queries of the test split and the generated irrelevant queries. Since FEVER is a fact-verification task, the irrelevant queries of the test split (derived from sentence claims) are entity-centric while the generated queries are more general or concept-focused (check \autoref{tab:fever} for examples of generated queries). This is because the model is copying the style of MSMARCO exemplars used in the QGen prompt. \textsc{Generate-Relevant-Only} is probably not affected because the irrelevant query-document examples are derived from a retriever i.e. for each generated relevant query the retriever retrieves the irrelevant document, and since most generated relevant queries are already entity-centric, the irrelevant query-document examples align well with the test split distribution.
Another reason for the poor performance of other QGen models could also be attributed to the fact that many examples (8.6\% $\sim$ 8k)  have document context of less than 20 words, which might not provide sufficient context for the model to generate good quality irrelevant queries. 
However, for the Touch\'e task, from \autoref{tab:touche} we see that this is not an issue. 
The generated queries from \textsc{Generate-Pairwise} are indeed not as relevant as the generated relevant queries.
Below, we conduct ablation experiments to better understand the efficacy of pairwise generation.

\begin{table}[t]
\resizebox{\columnwidth}{!}{
\begin{tabular}{lll}
\toprule
\textbf{QGen Prompt Source} & \textbf{QGen Model} & \textbf{NDCG@10} \\
\midrule
MS-MARCO & Label-Conditioned & 40.05 \\
         & Generate-Pairwise & 47.69 \\
\midrule 
FIQA     & Label-Conditioned & 45.44 \\
         & Generate-Pairwise & \textbf{56.45} \\
\bottomrule
\end{tabular}
}
\caption{We compare the effect of using task-specific exemplars (FIQA, in this case) in the QGen prompt over the generic MS-MARCO exemplars, and report the NDCG@10 metric for two QGen models.
}
\label{tab:fiqa}
\end{table}
\paragraph{Do the trends hold if task-specific exemplars are used in the QGen prompt?} \textbf{Yes!}
As mentioned above, we chose to use the same set of exemplars from MS-MARCO to generate queries for all tasks, while Promptagator use 8 task-specific exemplars.
Therefore, for one of the tasks, FIQA-2018, we re-run two QGen models, namely \textsc{Label-Conditioned} and \textsc{Generate-Pairwise}, but this time constructing all prompts using 10 exemplars from the training corpus of FIQA-2018.
From \autoref{tab:fiqa}, we find for both the QGen models using task-specific exemplars leads to a gain in the performance (+5 points for \textsc{Label-Conditioned} model and +8 points for \textsc{Generate-Pairwise}).
This is expected given that the LLM is now given in-domain knowledge to start from, however, with this ablation we wanted to highlight that the gains of the pairwise approach still hold.


\begin{table*}
\resizebox{\textwidth}{!}{
\begin{tabular}{lllllll}
\toprule
\textbf{Model Size} & \textbf{Type} & \textbf{QGen Model} &\textbf{Touch\'e} & \textbf{DBPedia} & \textbf{Climate-Fever} & \textbf{FIQA} \\
\midrule
XXL    & Baseline & Generate-Relevant-Only & 0.6512 & 0.5471	& 0.4704	& 0.6027 \\
    & & Label-Conditioned & 0.7466	& 0.5152	& 0.4888	& 0.4005 \\
& Ours & Generate-Pairwise & \textbf{0.7881}	& \textbf{0.5631}	& \textbf{0.5562}	& 0.4769 \\
& & Iterative-Pairwise & 0.7762	& 0.5557	& 0.4972	& \textbf{0.6584} \\
\midrule
XL   & Baseline & Generate-Relevant-Only & 	0.6646	& 0.5153	& 0.4011	& 0.4636 \\
   & & Label-Conditioned & 0.7466	& 0.5152	& 0.4888	& 0.4005 \\
   & Ours & Generate-Pairwise & 0.\textbf{7958}	& 0.\textbf{4942}	& \textbf{0.5026}	& 0.3604 \\
   & & Iterative-Pairwise & 0.7903	& 0.5284	& 0.4667	& \textbf{0.5099} \\
\midrule
L & Baseline & Generate-Relevant-Only & 0.6793	& 0.5102	& 0.3959	& 0.4093 \\
& & Label-Conditioned & 0.7835	& 0.3573	& 0.4033	& 0.2288 \\
& Ours & Generate-Pairwise & \textbf{0.8249}	& 0.4871	& \textbf{0.4278}	& 0.2684 \\
& & Iterative-Pairwise & 	0.8070 &	\textbf{0.4927} &	0.4200	& \textbf{0.4568} \\
\bottomrule
\end{tabular}
}
\caption{Comparing the performance of the downstream models trained on the generated queries across different sizes. Note here the generated queries for each downstream model is still from the PaLM-2 (M) model, we vary the downstream model size and measure the NDCG@10 (higher the better).}
\label{tab:results_models}
\end{table*}
\paragraph{Do the trends hold for smaller downstream models?} \textbf{Yes!}
Oftentimes, for many real-world applications, inference costs with respect to time, compute and latency, matter more than the fine-tuning, since inference needs to run several times and often on large set of documents.\footnote{\url{https://shorturl.at/jkAUZ}}
Therefore, through this ablation study, we aim to examine the performance of proposed QGen models across smaller downstream models.
In \autoref{tab:results_models}, we compare the downstream performance of three mt5-* models, namely, mt5-Large (1.2B), mt5-XL (3.7B), and mt5-XXL (13B) on four tasks.\footnote{\url{https://github.com/google-research/multilingual-t5}}
We find that the trends across all model sizes are consistent, with the proposed pairwise query generation models outperforming other QGen models.
In fact, for TREC-COVID the smallest model is the best performing.
This is an important result as it suggests that for inference settings where there is not sufficient compute or the requirement is to, say rank, millions of documents, smaller models also benefit from the queries generated using the proposed QGen models.

For all the above experiments, we generated queries for binary relevance labels.
However, some tasks have nuanced relevance judgements, for instance, TREC datasets are often annotated on 4-point likert scale.
Similarly, \citet{reddy2022shopping} released a nuanced shopping relevance dataset annotated on a 4-point scale, for improving product search.
This makes us wonder \emph{can we generate queries for such nuanced relevance labels using the above QGen approaches?}.

\section{Case Study: QGen for Fine-grained Relevance Prediction} \label{sec:finegrained}
To answer the above question, we follow \citet{chaudhary2023exploring} and conduct a limited study where we use ESCI \cite{reddy2022shopping}, a shopping relevance dataset, to generate queries for four relevance labels, and apply the generated queries to new products from WANDS dataset \cite{wands} and measure the downstream ranking performance.

\subsection{Data Setup}
\paragraph{ESCI} comprises of 2.6 million human-labeled query-product relevance judgements, derived from Amazon Search.
The query-product pairs are rated using four relevance labels: \emph{exact} when the product exactly matches the query specifications, \emph{substitute} when the product does not match exactly all requirements but could be used as a valid substitute, \emph{complement} when the product could be used in combination with the requested product, and \emph{irrelevant}.
ESCI is a multilingual dataset, including query-products for English, Spanish, and Japanese.
We use the English portion of the training data for our QGen experiments.

\paragraph{WANDS} is also a shopping relevance dataset, but derived from Wayfair Search, comprising of 233,448 human-annotated query-product relevance judgements, with 42,994 unique products.
The relevance labels are three-way, namely, \emph{exact-match} when the product exact satisfies the query requirements, \emph{partial-match} when the product is relevant to the query with respect to the main entity but differs in the exact modifiers requested, and \emph{irrelevant}.
We use the entire dataset as our test split, and all of the 42k products for query generation.
\begin{table*}
\resizebox{\textwidth}{!}{
\begin{tabular}{llllll}
\toprule
& \textbf{QGen Model} & \textbf{Num Synthetic Data} & \textbf{NDCG@5} & \textbf{NDCG@10} & \textbf{NDCG@20} \\
\midrule
Skyline & Transfer Learning* & 0 & 	\underline{0.8902} & \underline{0.8927} & \underline{0.8987} \\

\midrule
Baseline & FINETUNE-BASED-LABELCOND & 117179 & 0.8470 & 0.8553 & 0.8661 \\
& \cite{chaudhary2023exploring} & & & & \\
& Generate-Relevant-Only & 164185 & 0.5906 & 0.6112 & 0.6366 \\
& Label-Conditioned & 120856 & 0.8757  & 0.8779 &	0.8872 \\
\midrule
Ours & Generate-Pairwise & 168458 & \textbf{0.8835} & \textbf{0.8882}  & \textbf{0.8966 }\\
& Generate-AllLabels & 62036 &  0.8645  & 0.8700 & 0.8808\\
\bottomrule
\end{tabular}
}
\caption{We compare the NDCG scores of the different downstream models trained using data from the respective QGen models for the task of fine-grained relevance prediction on WANDS dataset.
We also include the best performing QGen model from \citet{chaudhary2023exploring}, i.e. \textsc{FINETUNE-BASED-LABELCOND} which is an mt5-XXL model fine-tuned for the query generation task, using relevance labels and documents as conditioning context. *The skyline results are used from \citet{chaudhary2023exploring}. 
}
\label{tab:wands}
\end{table*}
\subsection{Model Setup}
In this setting, we construct  prompts for all the QGen models using examples from ESCI.
The exact prompt formats are outlined in \autoref{tab:finegrainedprompts}.

\paragraph{Generate-Relevant-Only}
We use the same setting as described in \autoref{sec:baseline}, with the difference that instead of selecting exemplars from MS-MARCO, we select 10 query-product pairs from ESCI with the relevance label as \emph{exact}.
Next, we generate 2 relevant queries for each WANDS product, filter the generated queries and run the  retriever (\autoref{sec:baseline}) to retrieve irrelevant query-product pairs, and train a downstream model.

\paragraph{Label-Conditioned}
We select 10 query-product exemplars, ensuring at least two examples for each of the four relevance labels are included.
Next, we generate queries for each WANDS product and each of the four relevance labels.
Since we use beam size as 2, we expect to generate 8 queries for each product.
Next, we prompt the model to generate the relevance label, and remove those queries whose predicted label does not match the label using which it was generated.\footnote{Earlier, we had use the LLM's scoring mode for filtration, in this setting, we directly make the model generate the relevance label for a faster inference. For scoring across four labels, the LLM would have to make four calls for each input, which is expensive when the generated queries are O(100k)}.

\paragraph{Generate-Pairwise}
Earlier, we prompted the model to generate a relevant query followed by an irrelevant query.
Given that in this case there are four relevance labels, we adapt the task of pairwise generation to also take into account for which two labels the queries need to be generated.
Specifically, we prompt the model with the product and an instruction outlining for which two of the four labels the model should generate the queries (\autoref{tab:finegrainedprompts}).\footnote{Since $P(4,2)\!-\!$ 12 label permutations are possible, which would lead to $>10$ exemplars, we prioritize including those label combinations where the first label's relevance is higher than the second label's relevance, for example, pairs such as exact-substitute, exact-irrelevant.
We do include two exemplars where the the first query to be generated is for irrelevant label, ensuring that the model has seen at least one example for all four labels.}
Next, to generate new queries for each WANDS product, we create four label-combinations, namely, exact-complement, complement-exact, substitute-irrelevant, and irrelevant-substitute.\footnote{There were two reasons for these particular combinations: 1) these combinations ensure that for each product we generate both query1 and query2 for each relevance label, and 2) by skipping adjacent labels we hope to improve the generated output quality as our initial experiments showed that the adjacent labels often do not have clear separation boundary.}
As before, for each setting we generate two queries and then filter the queries whose predicted label does not match the label for which it was generated.

\paragraph{Generate-AllLabels}
We also extend the \textsc{Generate-Pairwise} model to generate queries for all labels at the same time.
Similar to how we generate relevant query first and irrelevant query second, in this setting, we prompt the model to also generate queries in the decreasing order of relevance.
After generating queries for each WANDS product, we filter the queries as before.

\subsection{Results}
In  \autoref{tab:wands} we find that among all QGen models, \textsc{Generate-Pairwise} performs the best.
\citet{chaudhary2023exploring} found that a model (mt5-XXL) fine-tuned for query-generation using labels as conditioning context outperformed the prompt-based \textsc{Label-Conditioned} QGen model.
Interestingly, we find here that our prompt-based \textsc{Label-Conditioned} model already outperforms \citet{chaudhary2023exploring}'s best model, which was fine-tuned on millions of ESCI training examples, highlighting that PaLM-2 (M) model is a strong foundational model even when used in a few-shot prompt fashion.
Within our proposed pairwise generation, we find that requiring the model to generate queries for all four labels (\textsc{Generate-AllLabels}) leads to only 62k training examples as opposed to 168k examples generated by restricting the model to generate  queries for only two labels.
This is because nearly 46\% of the LLM outputs were incorrectly formatted due to which the queries could not be parsed.\footnote{Details in \autoref{sec:app_wands}}
This is probably why the performance of \textsc{Generate-AllLabels} is lower than \textsc{Generate-Pairwise}.
From \autoref{tab:wands_outputs}, we see that the generated queries for \textsc{Generate-Pairwise} are indeed in the decreasing order of relevance, while both the relevant and not-relevant queries for \textsc{Generate-Relevant-Only} are in fact relevant to the product, which probably explains the poor performance of the latter model.

\section{Related Work}
LLMs have been extensively explored for synthetic text generation across different tasks, such as, Structured Prediction \cite{chen2023mixture}, Text Classification \cite{gao2022self, meng2022generating, gupta2023targen, wang-etal-2023-self-instruct, yu-etal-2023-regen}, Information Retrieval \cite{dai2022promptagator, bonifacio2022inpars}, to name a few.
\paragraph{LLMs for Query Generation}
As mentioned above, Promptagator \cite{dai2022promptagator} and InPars \cite{bonifacio2022inpars} both generate relevant queries, by prompting the model with relevant query-document examples.
InPars also propose a second prompt variant, called as GBQ (Guided By Bad Questions), where the LLM is shown a bad question or an irrelevant question for every relevant query-document example, similar to how we show irrelevant query for every relevant query-document example.
However, InPars discard the generated irrelevant questions for a new document and use a retriever to retrieve irrelevant query-document examples.
Another point of difference between Promptagator, InPars and our work is the type of downstream training.
Because our main focus is to compare different QGen approaches, we use a simple pointwise objective function i.e. pose the task as binary classification, while Promptagator and InPars use ranking-specific objective functions.

\paragraph{Task-Specific Prompting for Synthetic Text Generation}
The most common prompting strategy for text generation has been to use seed exemplars where the prompt is guided by a label-description (for example, Self-Instruct\cite{wang-etal-2023-self-instruct} and AttrPrompt \cite{yu2023large} for text classification tasks).
Specifically, AttrPrompt find that simply conditioned on class often leads to lack of diversity in generated examples and propose to condition on other attributes such as length and style.
More recently, \citet{gupta2023targen} propose TarGEN to generate synthetic data for text classification tasks, without using seed exemplars. Instead, TarGEN aim to generate diverse examples by incorporating task-specific elements (e.g. linguistic information about the task) in the text generation process.
They first generate these task-specific elements (these are not actual exemplars but intermediate hints useful for the task) and 
similar to \citet{chaudhary2023exploring} they also condition their example generation on the target label.
The generated examples are run through a LLM with a  task-specific prompt to remove examples, similar to \autoref{sec:filter}.
The above works suggest that adding task-specific attributes is a useful prompting strategy in addition to the final task labels, which aligns well with our finding as well, where using a ranking-based conditioning (e.g. \textsc{Generate-Pairwise} and \textsc{Iterative-Pairwise}) is helpful for ranking tasks.

\paragraph{LLMs for Relevance Judgements}
Complementary to our work, LLMs have also been directly used for relevance judgements or ranking documents \cite{thomas2023large, qin2023large, zhuang2023beyond, faggioli2023perspectives} via prompting.
\citet{qin2023large} propose the pairwise ranking prompting (PRP) that uses a query and a pair of documents as prompt to the LLMs for ranking. Specifically, they find that instead of asking the model to rank all documents in a list, providing documents in pair simplifies the ranking problem.
In addition to using binary relevance labels (relevant and irrelevant), some existing works \cite{zhuang2023beyond, faggioli2023perspectives} also prompt LLMs with intermediate relevance labels (e.g. partially relevant) to improvie fine-grained relevance prediction.

\section{Next Steps}
In this work, we have shown that by aligning the query generation task with the downstream ranking task, we are able to generate synthetic data useful to the downstream task.
In all the tasks, the query generation is conditioned on the document or a product context.
However, there are tasks within BEIR such as the Duplicate-Question Retrieval task where the task is retrieving a duplicate or similar question.
Given there is no conditioning context, it will be interesting to check how to adapt our best performing pairwise generation for such a task. 
As we saw in \autoref{tab:fever}, the generated irrelevant queries for FEVER were not grounded to the entities in the document.
Therefore, we plan to explore approaches to control the LLM generation along multiple aspects such as relevance, diversity, task-specific information.

\section{Limitations}
In this paper, we evaluate the generated query quality by running the downstream model for each QGen approach, which amounts to running a large set of experiments across different tasks resulting in computation and time cost.
We believe that having evaluation techniques which could simulate the downstream performance without requiring to run the downstream models every time, will save a lot of compute and time for researchers exploring the synthetic text generation space.

\section*{Acknowledgements}
We would like to thank Krishna Srinivasan, Honglei Zhuang, and Kazuma Hashimoto for their valuable suggestions and inputs throughout the project.

\bibliography{anthology,custom}

\clearpage
\appendix

\section{Appendix}

\subsection{Target Task Source} \label{sec:app_data}
Since some of the datasets used in BEIR are constantly updated, to maintain reproducibility and consistency, we download the datasets provided by BEIR authors hosted on TFDS\footnote{\url{https://www.tensorflow.org/datasets/catalog/beir}}.
Although, BEIR has a total of 18 datasets, with the exception of three categories Tweet-Retrieval, News Retrieval, Duplicate-Question Retrieval and Citation Prediction, we include at least one dataset from the other five task categories. Due to proprietary issues, datasets for Tweet-Retrieval and News Retrieval were absent from the TFDS library. We decided to exclude Duplicate-Question Retrieval as the task focused on query-query similarity which means there is no document context as such to generate the query from. Hence, for this work, we decided to focus on tasks having a provided document context.

\begin{table*}
\resizebox{\textwidth}{!}{
\begin{tabular}{l}
\midrule
Generate-Relevant-Only \\
\midrule
Given a product generate a query that exactly matches the product specifications: \\
\\

product: MOUNTAINTOP 40L Hiking Backpacks with Rain Cover for Women Men MOUNTAINTOP 40L Hiking Backpack,\\
A must have for hiking, camping, traveling and cycling, helps you reach the most epic views the world has to offer \\
query: mountaintop hiking pack \\
\\
product: {\texttt{new product}} \\
\midrule

Label-Conditioned \\
\midrule
Given a product and desired relevance label generate a query that is appropriate for that relevance label. \\
The four relevance labels are `Exact' which means that the item is relevant for the query,\\
and satisfies all the query specifications . `Substitute' means that the item is somewhat relevant, i.e., it fails \\
to fulfill some aspects of the query but the item can be used as a functional substitute. \\
`Complement'  means that the item does not fulfill the query, but could be used in combination with an exact item. \\
`Irrelevant'  means that the item is irrelevant, or it fails to fulfill a central aspect of the query. 
Some examples are: \\
\\

product: MOUNTAINTOP 40L Hiking Backpacks with Rain Cover for Women Men MOUNTAINTOP 40L Hiking Backpack,\\ A must have for hiking, camping, traveling and cycling, helps you reach the most epic views the world has to offer \\
label: Exact \\
query: mountaintop hiking pack \\
\\

product: Office Chair Ergonomic Cheap Desk Chair Mesh \\ Computer Chair Lumbar Support Modern Executive Adjustable Stool Rolling Swivel Chair for Back Pain, Black \\
label: Substitute \\
query: office chair without wheels or lift \\
\\
product: {\texttt{new product}} \\
label: {\texttt{Exact/Substitute/Complement/Irrelevant}} \\
\midrule
Generate-Pairwise \\
\midrule
Given a product and a desired relevance label, the task is to generate two unique query for each relevance label.\\
The four relevance labels are 'Exact' which means that the item is relevant for the query, and satisfies all the query specifications .
'Substitute' means that the \\item is somewhat relevant, i.e., it fails \\
to fulfill some aspects of the query but the item can be used as a functional substitute. \\
'Complement'  means that the item does not fulfill the query, but could be used in combination with an exact item. \\
'Irrelevant'  means that the item is irrelevant, or it fails to fulfill a central aspect of the query. \\
\\

product: MOUNTAINTOP 40L Hiking Backpacks with Rain Cover for Women Men MOUNTAINTOP 40L Hiking Backpack,\\ A must have for hiking, camping, traveling and cycling, helps you reach the most epic views the world has to offer\\
task: generate query1 for Exact and query2 for Substitute \\
query1: mountaintop hiking pack \\
query2: osprey jet 12 \\
\\

product: Office Chair Ergonomic Cheap Desk Chair Mesh \\ Computer Chair Lumbar Support Modern Executive Adjustable Stool Rolling Swivel Chair for Back Pain, Black \\
task: generate query1 for Substitute and query2 for Complement \\
query1: office chair without wheels or lift \\
query2: ergonimic foot stool \\
\\

product: {\texttt{new product}}\\
task: generate query1 for {\texttt{label-1}} and query2 for {\texttt{label-2}} \\
\midrule
Generate-AllLabels \\
\midrule
Given a product and a desired relevance label, the task is to generate a unique query for each relevance label.\\
The four relevance labels are `Exact' which means that the item is relevant for the query, and satisfies all the query specifications . `Substitute' means that the item \\is somewhat relevant, i.e., it fails \\to fulfill some aspects of the query but the item can be used as a functional substitute. \\
`Complement'  means that the item does not fulfill the query, but could be used in combination with an exact item. \\
`Irrelevant'  means that the item is irrelevant, or it fails to fulfill a central aspect of the query. \\
\\

product: MOUNTAINTOP 40L Hiking Backpacks with Rain Cover for Women Men MOUNTAINTOP 40L Hiking Backpack,\\ A must have for hiking, camping, traveling and cycling, helps you reach the most epic views the world has to offer \\
Label: Exact Query: mountaintop hiking pack \\
Label: Substitute Query: osprey jet 12 \\
Label: Complement Query: waterproof shoes hiking \\
Label: Irrelevant Query: mountaintop whitlow \\
\\
product: {\texttt{new product}} \\

\bottomrule
\end{tabular}
}
\caption{Prompt formats for the different QGen models used in \autoref{sec:finegrained}.
}
\label{tab:finegrainedprompts}
\end{table*}
\begin{table*}
\resizebox{\textwidth}{!}{
\begin{tabular}{l}
\midrule
Generate-Relevant-Only \\
\midrule
Given a passage from a web page, generate a search query for which the passage can be a perfect answer.\\
\\

passage: Premature Ventricular Contractions (PVCs, PVC) Medical Definition of Cardiac stress testing, exercise.\\
Cardiac stress testing, exercise: The exercise cardiac stress testing (EST) is the most widely used cardiac (heart) screening test.\\
The patient exercises on a treadmill according to a standardized protocol, with progressive increases in the speed \\
and elevation of the treadmill (typically changing at three-minute intervals).\\
query: what is cardiac testing in medical terms \\
\\

passage: {\texttt{new passage}} \\
query: {\texttt{relevant query}} \\
\midrule

Label-Conditioned \\
\midrule
Given a passage from a web page and a relevance label, generate a search query appropriate for that relevance level for that passage.\\
If the label is "relevant", the query should be such that the passage can be a perfect answer and if the label is "irrelevant" the query\\ should be such that the passage is not a perfect answer.\\
\\

passage: Premature Ventricular Contractions (PVCs, PVC) Medical Definition of Cardiac stress testing, exercise.\\ Cardiac stress testing, exercise: The exercise cardiac stress testing (EST) is the most widely used cardiac (heart) screening test.\\
The patient exercises on a treadmill according to a standardized protocol, with progressive increases in the speed\\ and elevation of the treadmill (typically changing at three-minute intervals).\\
label: relevant\\
query: what is cardiac testing in medical terms\\
\\

passage: Amazon Customer Service Whatever the issue, you're going to want to get in touch with Amazon's customer service department.\\
The easiest way to contact Amazon's customer service department is by using their toll-free phone number at 1-888-280-4331.\\
label: irrelevant\\
query: amex customer service phone number\\
\\

passage: {\texttt{new passage}} \\
label: {\texttt{relevant / irrelevant}} \\
query: {\texttt{relevant / irrelevant query}} \\
\midrule
Generate-Pairwise \\
\midrule
Given a passage from a web page, generate a search query for which the passage can be a perfect answer and\\
a search query for which the passage is not a perfect answer.\\
\\

passage: Premature Ventricular Contractions (PVCs, PVC) Medical Definition of Cardiac stress testing, exercise.\\ Cardiac stress testing, exercise: The exercise cardiac stress testing (EST) is the most widely used cardiac (heart) screening test.\\ The patient exercises on a treadmill according to a standardized protocol, with progressive increases in the speed\\ and elevation of the treadmill (typically changing at three-minute intervals).\\
query1: what is cardiac testing in medical terms\\
query2: how soon exercise after heart stent\\
\\

passage: Amazon Customer Service Whatever the issue, you're going to want to get in touch with Amazon's customer service department.\\ The easiest way to contact Amazon's customer service department is by using their toll-free phone number at 1-888-280-4331.\\
query1: what is amazon phone number customer service\\
query2: amex customer service phone number\\
\\

passage: {\texttt{new passage}}\\
query1: {\texttt{relevant query}}\\
query2: {\texttt{irrelevant query}} \\
\bottomrule
\end{tabular}
}
\caption{Prompt formats for the different QGen models used in \autoref{sec:setup}.
}
\label{tab:prompt_formats}
\end{table*}

\subsection{QGen Data Statistics}
We report the number of queries generated for each step of the QGen process for all models in \autoref{tab:relevant_qgen} (\textsc{Generate-Relevant-Only}), \autoref{tab:labelcond_qgen} (\textsc{Label-Conditioned}), \autoref{tab:pairwise_qgen} (\textsc{Generate-Pairwise}), and \autoref{tab:iterative_qgen} (\textsc{Iterative-Pairwise}).

\subsection{Invalid Prompt Outputs} \label{sec:app_wands}
LLMs often generate invalid or useless outputs \cite{qin2023large}.
For the QGen experiments on BEIR datasets, where queries were generated for binary relevance labels, we find from \autoref{tab:pairwise_qgen}, \autoref{tab:relevant_qgen}, \autoref{tab:labelcond_qgen}, and \autoref{tab:iterative_qgen} that the percentage of such invalid queries generated is nearly 100\% for \textsc{Generate-Relevant-Only} and \textsc{Label-Conditioned}, while for \textsc{Generate-Pairwise} and \textsc{Iterative-Pairwise} it drops to 85\% and 87\% respectively.
Some common reasons for such invalid outputs are lack of `query' or `label' prefix which causes incorrect parsing.
For QGen approaches where only one query output is expected (i.e. \textsc{Generate-Relevant-Only} and \textsc{Label-Conditioned}) such issues are not that common but for the other two approaches where two queries need to be generated such errors are observed more. 
We observe this issue more for \textsc{Generate-Labels} approach used in the fine-grained relevance prediction case study, where the model is expected to generate four queries for four labels.
In particular, we find 46\% of generated queries are invalid or useless which dramatically reduces the synthetic data pool.
We find that most of these are deemed invalid because the LLM starts generating garbage and does not adhere to the task instruction, nearly all of the 46\% invalid queries are outputs where the LLM generates information about a new product, its title and description which is typically provided as input.

\begin{table*}
\resizebox{\textwidth}{!}{
\begin{tabular}{lll|ll|lll|lll}
\toprule
& \multicolumn{2}{l|}{QGen Inputs} & \multicolumn{2}{l|}{Process QGen Outputs} & \multicolumn{3}{l}{Training Data} & & \\
\textbf{Dataset} & \textbf{Prompt} & \textbf{Requested} & \textbf{Valid Query} & \textbf{Filtered Query} & \textbf{Train} & \textbf{Irrelevant} & \textbf{Relevant} & \textbf{\% Valid} &  \textbf{\% Valid} & \textbf{Irrelevant / Relevant}\\
& \textbf{Inputs} & \textbf{Queries} & \textbf{Outputs} & \textbf{Outputs} & \textbf{Examples} & \textbf{Examples} & \textbf{Examples} & \textbf{Queries} & \textbf{Examples}  \\
\midrule
trec-covid & 128821 & 257642 & 457072 & 257204 & 257204 & 85385 & 171819 & 0.89 & 0.5 & 0.5\\
touche & 50001 & 100002 & 168370 & 88295 & 88295 & 67495 & 20800 & 0.84 & 0.44 & 3.24\\
dbpedia & 50001 & 100002 & 167317 & 93655 & 93655 & 73587 & 20068 & 0.84 & 0.47 & 3.67\\
climate-fever & 50001 & 100002 & 166364 & 88872 & 88872 & 76488 & 12384 & 0.84 & 0.44 & 6.18\\
fiqa & 57013 & 114026 & 199491 & 109593 & 109593 & 72296 & 37297 & 0.87 & 0.48 & 1.94\\
fever & 50001 & 100002 & 166047 & 88696 & 88696 & 76864 & 11832 & 0.83 & 0.44 & 6.5\\
\bottomrule
\end{tabular}
}
\caption{We report the data statistics for each step of \textsc{Generate-Pairwise} model for all datasets. c
}
\label{tab:pairwise_qgen}
\end{table*}
\begin{table*}
\resizebox{\textwidth}{!}{
\begin{tabular}{lll|ll|lll|lll}
\toprule
& \multicolumn{2}{l|}{QGen Inputs} & \multicolumn{2}{l|}{Process QGen Outputs} & \multicolumn{3}{l}{Training Data} & & \\
\textbf{Dataset} & \textbf{Prompt} & \textbf{Requested} & \textbf{Valid Query} & \textbf{Filtered Query} & \textbf{Train} & \textbf{Irrelevant} & \textbf{Relevant} & \textbf{\% Valid} & \textbf{\% Valid} & \textbf{Irrelevant / Relevant}\\
& \textbf{Inputs} & \textbf{Queries} & \textbf{Outputs} & \textbf{Outputs} & \textbf{Examples} & \textbf{Examples} & \textbf{Examples} & \textbf{Queries} & \textbf{Examples} \\
\midrule
trec-covid & 128821 & 257642 & 257497 & 197995 & 360465 & 162470 & 197995 & 0.99 & 0.77 & 0.82\\
touche & 50001 & 100002 & 96003 & 25598 & 51196 & 25598 & 25598 & 0.96 & 0.26 & 1.0\\
dbpedia & 50001 & 100002 & 100001 & 24064 & 46198 & 22134 & 24064 & 0.99 & 0.24 & 0.92\\
climate-fever & 50001 & 100002 & 99978 & 14171 & 27208 & 13037 & 14171 & 0.99 & 0.14 & 0.92\\
fiqa & 57013 & 114026 & 113536 & 22182 & 43814 & 21002 & 22182 & 0.99 & 0.19 & 0.95\\
fever & 50001 & 100002 & 99970 & 13793 & 26585 & 12792 & 13793 & 0.99 & 0.14 & 0.93\\

\bottomrule
\end{tabular}
}
\caption{We report the data statistics for each step of \textsc{Generate-Relevant-Only} model for all datasets. The \% of valid queries refers to the percentage of valid queries after the query generation step while the \% of valid examples is after the filtration step.
}
\label{tab:relevant_qgen}
\end{table*}
\begin{table*}
\resizebox{\textwidth}{!}{
\begin{tabular}{lll|ll|lll|lll}
\toprule
& \multicolumn{2}{l|}{QGen Inputs} & \multicolumn{2}{l|}{Process QGen Outputs} & \multicolumn{3}{l}{Training Data} & & \\
\textbf{Dataset} & \textbf{Prompt} & \textbf{Requested} & \textbf{Valid Query} & \textbf{Filtered Query} & \textbf{Train} & \textbf{Irrelevant} & \textbf{Relevant} & \textbf{\% Valid} &  \textbf{\% Valid} & \textbf{Irrelevant / Relevant}\\
& \textbf{Inputs} & \textbf{Queries} & \textbf{Outputs} & \textbf{Outputs} & \textbf{Examples} & \textbf{Examples} &  \textbf{Examples} & \textbf{Queries} & \textbf{Examples} \\
\midrule
trec-covid & 257642 & 515284 & 515258 & 305238 & 305238 & 117693 & 190485 & 0.99 & 0.59 & 0.62\\
touche & 100002 & 200004 & 199992 & 103542 & 103542 & 78561 & 26697 & 0.99 & 0.52 & 2.94\\
dbpedia & 100002 & 200004 & 200000 & 105665 & 105665 & 85981 & 22129 & 0.99 & 0.53 & 3.89\\
climate-fever & 100002 & 200004 & 199987 & 103043 & 103043 & 91465 & 13558 & 0.99 & 0.52 & 6.75\\
fiqa & 114026 & 228052 & 228052 & 118615 & 118615 & 71905 & 47289 & 0.99 & 0.52 & 1.52\\
fever & 100002 & 200004 & 199986 & 102924 & 102924 & 91879 & 12970 & 0.99 & 0.51 & 7.08\\
\bottomrule
\end{tabular}
}
\caption{We report the data statistics for each step of \textsc{Label-Conditioned} model for all datasets.  The \% of valid queries refers to the percentage of valid queries after the query generation step while the \% of valid examples is after the filtration step.
}
\label{tab:labelcond_qgen}
\end{table*}
\begin{table*}
\resizebox{\textwidth}{!}{
\begin{tabular}{llll|l|ll|lll}
\toprule
& \multicolumn{3}{l|}{QGen Inputs} & \multicolumn{1}{l|}{Process QGen Outputs} & \multicolumn{3}{l}{Training Data} & & \\
\textbf{Dataset} & \textbf{Prompt} & \textbf{Requested} & \textbf{Relevant Query} & \textbf{Irrelevant}  & \textbf{Train} & \textbf{Irrelevant} & \textbf{\% Valid} & \textbf{\% Valid} & \textbf{Irrelevant / Relevant}\\
& \textbf{Inputs} & \textbf{Queries} & \textbf{Outputs} & \textbf{Query Requested} & \textbf{Outputs} & \textbf{Examples} & \textbf{Queries} & \textbf{Examples}  \\
\midrule
trec-covid & 128821 & 257642 & 197995 & 404323 & 287490 & 485485 & 0.71 & 0.94 & 1.45\\
touche & 50001 & 100002 & 25598 & 50524 & 39379 & 64977  & 0.77 & 0.32 & 1.54\\
dbpedia & 50001 & 100002 & 24064 & 47980 & 41748 & 65812 & 0.87 & 0.33 & 1.73\\
climate-fever & 50001 & 100002 & 14171 & 28218 & 28555 & 42726 & 1.0 & 0.21 & 2.02\\
fiqa & 57013 & 114026 & 22182 & 44018 & 38749 & 60931 & 0.88 & 0.27 & 1.75\\
fever & 50001 & 100002 & 13793 & 27485 & 27877 & 41670 & 1.0 & 0.21 & 2.02\\
\bottomrule
\end{tabular}
}
\caption{We report the data statistics for each step of \textsc{Iterative-Pairwise} model for all datasets.  The \% of valid queries refers to the percentage of valid queries after the query generation step while the \% of valid examples is after the filtration step.
}
\label{tab:iterative_qgen}
\end{table*}
\begin{table*}
\resizebox{\textwidth}{!}{
\begin{tabular}{l|l}
\midrule
Because political parties control primary elections, & I would not need to explain what \\
party bosses are given substantial power & chopsticks are. I assume the position that\\
and authority in making decisions that &   a set of chopsticks is a superior \\
lead to the election of the  next president.This is undemocratic.&  eating utensil to a fork.\\
\midrule
\multicolumn{2}{l}{Generate-Relevant-Only}\\
\textbf{relevant}: why primary elections are undemocratic & why are chopsticks superior to forks\\
\textbf{irrelevant}: how do superdelegates undermine the democratic party & chop sticks are better than forks\\

\midrule
\multicolumn{2}{l}{Label-Conditioned} \\
\textbf{relevant}: why are political parties undemocratic & why is chopsticks better than fork?\\
\textbf{irrelevant}: why are political parties bad & what are the best eating utensils\\

\midrule
\multicolumn{2}{l}{Generate-Pairwise} \\
\textbf{relevant}: why are primaries undemocratic & are chopsticks superior to a fork\\
\textbf{irrelevant}: why is voting not democratic & why chopsticks are better than knife\\

\midrule
\multicolumn{2}{l}{Iterative-Pairwise} \\
\textbf{relevant}: why primary elections are undemocratic & why are chopsticks superior to forks\\
\textbf{irrelevant}: why primary elections & what is the best utensil for eating\\

\bottomrule
\end{tabular}
}
\caption{We compare the generated queries across different QGen models for the Touch\'e dataset. The first row denotes the passage snippets used for the query generation. Note for the \textsc{Generate-Relevant-Only} the irrelevant queries are not specifically generated rather the generated relevant queries are paired with retrieved documents.
}
\label{tab:touche}
\end{table*}

\begin{table*}
\resizebox{\textwidth}{!}{
\begin{tabular}{l|l}
\midrule
The Painter and The Buyer is a 1565 pen and ink on & Mateuți is a commune in Rezina District, Moldova.\\
brown paper painting by Flemish artist Pieter Bruegel  & It is composed of a single village , Mateuți .\\
the Elder. The alternative title is The Artist  &    \\
and The Connoisseur.The painter is thought &  \\
to be a self-portrait of Pieter Bruegel the Elder. & \\
\midrule
\multicolumn{2}{l}{Generate-Relevant-Only}\\
\textbf{relevant}: what is the painter and the buyer painting & what is mateuti\\
\textbf{irrelevant}: who is drew halfmann & what is dancu district in moldova\\
\midrule
\multicolumn{2}{l}{Label-Conditioned} \\
\textbf{relevant}: who is painter in painter and buyer & what is mateuti\\
\textbf{irrelevant}: what is the difference between a painting and a picture & how to read vitamin labels\\

\midrule
\multicolumn{2}{l}{Generate-Pairwise} \\
\textbf{relevant}: is painter and the buyer a 1565 pen and ink on & what is mateuti\\
          brown paper painting by Flemish artist Pieter Bruegel the Elder & \\
\textbf{irrelevant}: what is the name of the painter & what is the most common name in the world\\

\midrule
\multicolumn{2}{l}{Iterative-Pairwise} \\
\textbf{relevant}: what is the painter and the buyer painting & what is mateuti\\
\textbf{irrelevant}: what is the painter and the buyer about & what is oxegen\\

\bottomrule
\end{tabular}
}
\caption{We compare the generated queries across different QGen models for the FEVER dataset. The first row denotes the passage used for the query generation. Note for the \textsc{Generate-Relevant-Only} the \textbf{relevant} queries are not specifically generated rather the generated relevant queries are paired with retrieved documents.
}
\label{tab:fever}
\end{table*}
\begin{table*}
\resizebox{\textwidth}{!}{
\begin{tabular}{l|l}
\toprule
rangeworthy 67 '' genuine leather tuxedo arm loveseat 
& stainless steel modern pull 3 3/4 '' center to centerfinger pull  
\\
\midrule
\multicolumn{2}{l}{Generate-Relevant-Only}\\
\textbf{exact}: leather sofa for living room & stainless steel modern pull \\
\textbf{not-exact}: 67 '' leather loveseat &  stainless steel pull  \\

\midrule
\multicolumn{2}{l}{Label-Conditioned} \\
\textbf{exact}: leather couch & stainless steel pull  \\
\textbf{substitute}: - & - \\
\textbf{complement}: leather sectional for small space & best kitchen cabinet hardware  \\
\textbf{irrelevant}: baby girl shoes & baby crib  \\
\midrule

\multicolumn{2}{l}{Generate-Pairwise}\\
\textbf{exact}: leather sectional furniture & black stainless steel door pulls   \\
\textbf{substitute}: red leather lounge chair & stainless steel door knob  \\
\textbf{complement}: leather recliner for senior & -\\
\textbf{irrelevant}: leather chair for gaming & 4 in 1 baby car seat  \\
\midrule
\multicolumn{2}{l}{Generate-AllLabels} \\
\textbf{exact}: tuxedo arm loveseat & stainless steel modern pull 3 3/4 '' center to centerfinger pull \\
\textbf{substitute}: leather couch 70 inches & black modern drawer pulls  \\
\textbf{complement}: - & - \\
\textbf{irrelevant}: - & black modern furniture  \\

\bottomrule
\end{tabular}
}
\caption{We compare the generated queries for different QGEn models for the WANDS dataset. Note, exemplars from ESCI are used for the query generation. The first row denotes the WANDS product used for the query generation.
}
\label{tab:wands_outputs}
\end{table*}



\end{document}